# Aging Decline in Basketball Career Trend Prediction Based on Machine Learning and LSTM Model

Yi-chen Yao[1]   Jerry Wang[1,2]   Yi-cheng Lai[1,2]   Lyn Chao-ling Chen[3, *]
Department of Management Information Systems[1]
Bachelor Program in Artificial Intelligence Applications[2]
Interdisciplinary Artificial Intelligence Center[3]
National Chengchi University

Taipei, Taiwan
lynchen@ntu.edu.tw

**ABSTRACT**

**The topic of aging decline on performance of NBA players has been discussed in this study. The autoencoder with K-means clustering machine learning method was adopted to career trend classification of NBA players, and the LSTM deep learning method was adopted in performance prediction of each NBA player. The dataset was collected from the basketball game data of veteran NBA players. The contribution of the work performed better than the other methods with generalization ability for evaluating various types of NBA career trend, and can be applied in different types of sports in the field of sport analytics.**

## I. INTRODUCTION

In the field of sports analytics, machine learning methods are being widely used to analyze player behavior and performance prediction for the strong capabilities in pattern recognition and career trend prediction, and the topic of aging decline on the performance of basketball game players were discussed in the study. Elite veteran National Basketball Association (NBA) players have different levels of decline in the later stages of their career that depends on their skill acquisition phase [1]. Therefore, aging decline and career trend differ between the NBA star players and the NBA role players, and the NBA star players have stable performance for the improvement of basketball skills. Hence, instead of using the same approach for all players, a two-stage framework proposed that autoencoder with K-means clustering were adopted to classify types of career trend, and the Long Short-Term Memory (LSTM) model was adopted to the clustering results combined with time series data for career trend prediction of NBA players.

This paper is structured as follows: Section 2 reviews the traditional statistics, machine leaning methods and deep learning methods in basketball game analytics. Section 3 outlines the methodology that introduces a two-stage framework for prediction of NBA career trend. Section 4 compares the model performance and discussions of the aging patterns and the performance of NBA players. Section 5 concludes the study.

## II. RELATED WORKS

Data mining has used in basketball analytics, and a PC-based application, Advanced Scout, has developed by International Business Machines Corporation (IBM) in the 1990s to discover patterns in NBA data [2]. Machine learning method are commonly used to predict the performance and to identify types of NBA players. Using Bayes classification and multivariate linear regression in data mining performed well in predictions of winner and the value of spread in NBA games [3]. Scoring has been found that is the most important factor for any NBA team, and also the favorable style of basketball fans from both regression and classification analysis in small dataset [4]. Using both random forest and multiple linear regression in score performance prediction of NBA players achieved high performance [5]. From the development of individualized models of NBA players, meta-model, random forest, Bayesian Ridge, Adaptive Boosting (AdaBoost), and Elastic Net performed well in Fantasy Points (FP) performance [6]. Furthermore, the NBA performance prediction helps to optimize the player strategies and team performance, and gradient boosting regressor performed well by integrating advanced hyperparameter tuning and feature selection [7]. Using 50 years of NBA seasonal data, random forest was adopted for Most Valuable Player (MVP) prediction across different NBA basketball effectively in the data with rapid change context [8]. Deep learning methods have applied for complex pattern recognition in basketball analysis of NBA players [9] [10]. Multi-Layer Perceptron (MLP) model has trained on 600 game parameters to predict team performance, and from the comparison results of the 1%, 2% and 3% change in performance found that the change was smaller for lower performing teams and slightly larger for better performing teams [11]. Considering both textual data and historical data in performance perdition, using interview transcripts before the NBA games, both text-based BiLSTM (LSTM-T) and BERT attention (BERT-A-T) performed well in performance prediction of NBA players, and the BERT-A-T model most associated to each prediction task [12]. For processing time series data of NBA SportVu tracking data, both the LSTM model and the bidirectional long short-term memory (BLSTM) combined with the mixture density network (MDN) can predict motion trajectories of basketballs accurately [13] [14].

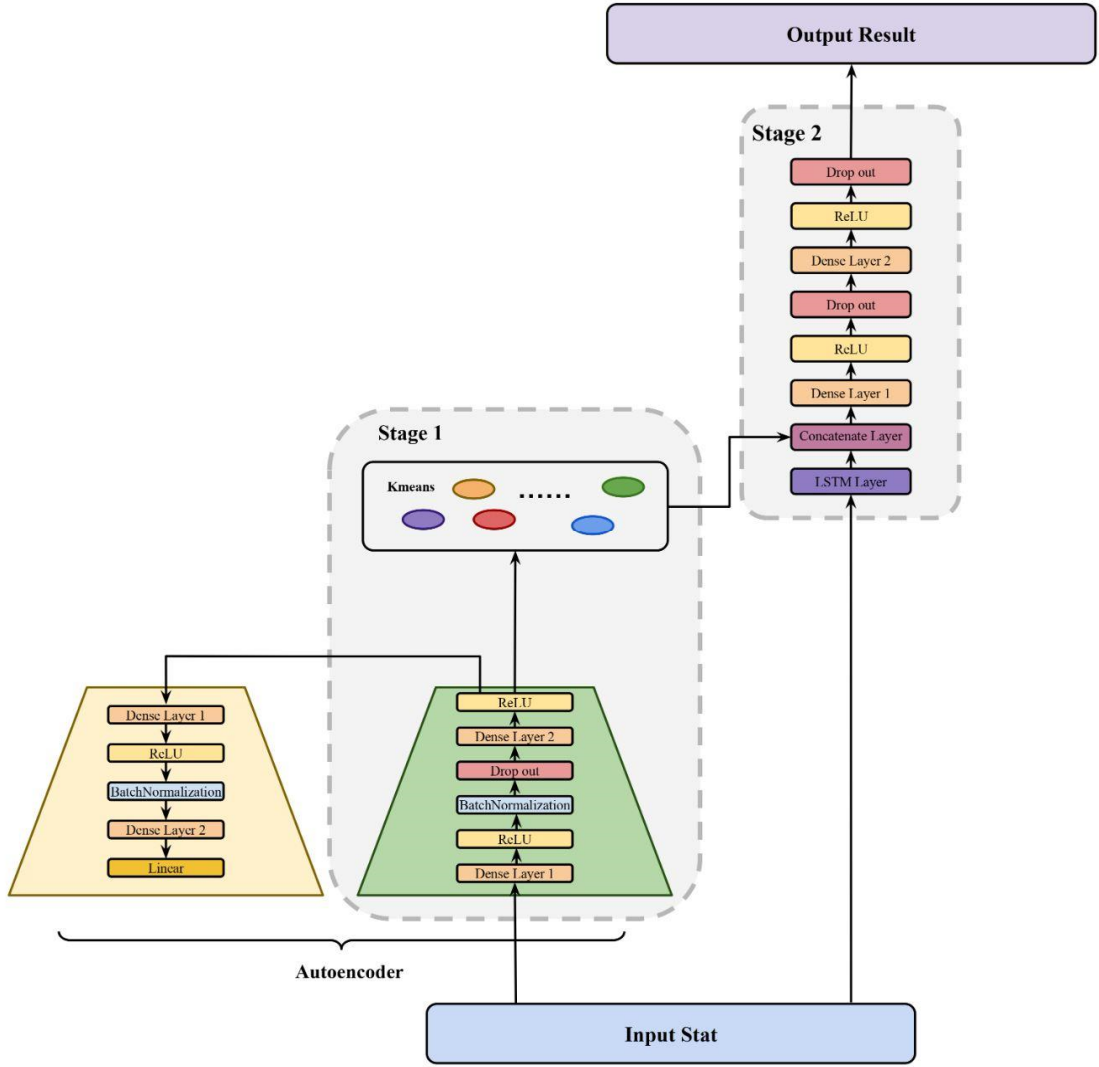

Figure 1. Two-stage framework of the proposed method.

Moreover, for identifying players types, the K-means clustering was adopted to evaluate NBA guards objectively using data of rebounds, assists and points [15]. In addition, a weighted network K-means clustering approach called Louvain method classified eight general player types across six years of NBA data [16]. Hence, in the study, autoencoder with K-means clustering and the LSTM model were adopted for career trend perdition of NBA players.

III. METHODOLOGY

A. Dataset

Using the dataset of NBA seasonal data from 1995 to 2023 with comprehensive historical data of 48 numerical features of each NBA player per season, and the dataset includes both basic box score statistics and advanced analytics metrics [17]. In the data preprocessing phase, data of at least 5 seasons with complete career trend between age 22 to 31 years old were selected for statistical reliability and avoiding overfitting, including all 48 numerical features such as including basic statistics of Points (PTS), Rebounds (REB) and Assists (AST), and advanced metrics of Box Plus/Minus (BPM), Player Efficiency Rating (PER) and True Shooting Percentage (TS). Career trend of each NBA player is structured as a 7-year developmental sequence (Age: 22 to 28 years old) with 48 features that pairs with a 3-year target sequence (Age: 29 to 31 years old, BPM) for career trend prediction. For filling the missing data, for example in TS feature used the median value from similar players, and in counting statistics used forward or backward data. Total of 177 NBA players with complete career trend in the dataset, consists of 141 NBA players in the training set and 36 NBA players in the test set, and no data both exists in the training set and the test set to avoid data leakage.

B. Problem Formulation

In the preliminary stage, using the data with complete career trends (Age: 22 to 32 years old), peak period, decline rate and overall career trend of different types of NBA players were evaluated. From the results of average BPM performance by age (Figure 2), elite players with high BPM performance have different career tends than the average players and below-average players with low BPM performance, and also explained that single approach may not applicable to identify types of career trend. Hence, the two-stage framework proposed in the study, stage 1 identifies the types of career trend, and stage 2 predicts the career trend of NBA players.

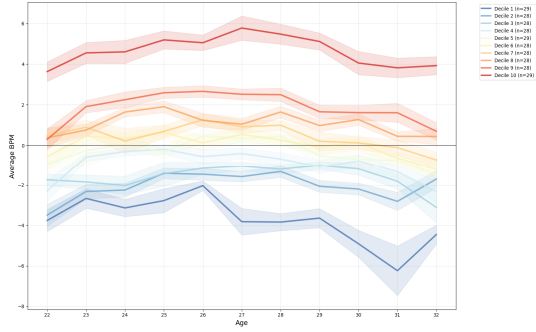

Figure 2. BPM performance of different types of NBA players by age.

TABLE I. SILHOUETTE SCORE OF CLUSTER NUMBER

| K Value | Silhouette Score |
|---|---|
| 2 | 0.16 |
| 3 | 0.15 |
| 4 | 0.12 |
| 5 | 0.11 |
| 6 | 0.10 |
| 7 | 0.10 |
| 8 | 0.09 |

The problem of NBA career trend prediction can be considered as a task of time series forecasting, and the work aims to predict future performance according to the types of the historical NBA career trend. Using data from the development period of NBA players (Age: 22 to 28 years old), the BPM performance of NBA players during the peak period (Age: 29 to 31 years old) can be predicted. Let $X = \{x_1, x_2, \ldots, x_7\}$ represents the input sequence, where each $x_i \in \mathbb{R}^{48}$ represents the feature vector at age $(21+i)$. The target sequence is $Y = \{y_1, y_2, y_3\}$, where $y_i$ represents BPM at age $(28+i)$, and the goal is to learn the mapping function $f$ in formula (1), and generates predictions $\hat{Y}$ to approximate the actual target values $Y$.

$$f: \mathbb{R}^{7 \times 48} \to \mathbb{R}^3 \text{ such that } \hat{Y} = f(X). \quad (1)$$

C. Two-stage Framework

The proposed method adopted clustering in the two-stage framework for NBA career trend prediction (Figure 1) (implemented in Python language), considering the differences of NBA career patterns for the different playing styles and paths in the development phase of NBA players, stage 1 identifies types of career trend through autoencoder of representation learning and K-means clustering, and the results of stage 1 as input of stage 2 combined with time series data predicts career trend though LSTM model. In stage 1, the autoencoder transforms the career trend sequences $X \in \mathbb{R}^{N \times 7 \times 48}$ into compact embeddings $Z \in \mathbb{R}^{N \times 64}$ to pair with the decoder for evaluating the best dimension of the encoder, and then the K-means clustering identifies types of NBA career trend to retrieve extra patterns. In stage 2, the LSTM model process both the career trend sequences and the corresponding cluster assignments for career trend prediction. For evaluation of model performance, the Mean Absolute Error (MAE) evaluates the prediction accuracy and provides interpretable error magnitude in BPM units with robustness, and the Coefficient of Determination ($R^2$) evaluates the ability of the model to capture patterns of the career trend data from the proportion of variance.

*1) Stage 1: Autoencoder with K-means Clustering for Career Trend Classification:* The input from historical career trend contains 48 features across 7 years with 336-dimensional vectors, the autoencoder was adopted to reduce the high dimensional data with complexity through an encoder-decoder structure before clustering. The encoder compresses the input $X \in \mathbb{R}^{336}$ through two dense layers for dimension reduction. The first layer reduces the 336 dimensions of input to 128 dimensions for narrowing down the features, and uses batch normalization and dropout (Rate=0.1) to prevent overfitting. The second layer reduces the 128 dimensions from first layer results to 64 dimensions with Rectified Linear Unit (ReLU) activation, and retrieves features of career trends of each NBA player in the $Z \in \mathbb{R}^{64}$ compressed representation. The decoder reverses the process to reconstruct the input that enables the encoder to learn useful features. If the 64 dimensions contain useful information for illustrating the career trend, the decoder successes to reconstruct the original data, otherwise, the reconstruction fails with high error. The reconstruction loss function is defined in formula (2) whether the encoder converges.

$$L = (1/N) \Sigma_{i=1}^{N} |X_i - Decoder(Encoder(X_i))|_2^2 \quad (2) \text{ [18]}$$

In the training phase, the Adam optimizer was used to ensure stable convergence and to prevent overfitting, and reduces the historical career trend data of each NBA player from 336 dimensions to 64 dimensions. After autoencoder training, K-means clustering was adopted to identify types of career trend. Using Silhouette score, the optimal K value can be determined (K=2) that high Silhouette score represents high performance of clustering (Table I), and two types of career trend were classified. This Silhouette score (0.16) reflects the limited sample size and similarity among the NBA players, but even weak clustering helps to generate promising prediction results. Hence, the types of career tend of each NBA player can be identified after clustering.

*2) LSTM Model for Carrer Trend Prediction:* LSTM model was adopted for career trend prediction, one input as the input of career trend sequences $X \in \mathbb{R}^{7 \times 48}$ in stage 1 with 48 features across 7 years (Age: 22 to 28 years old) that represents historical performance of each NBA player, and another input transforms the clustering results of stage 1 to one-hot encoded vectors $c \in \mathbb{R}^K$, where K is optimal value of clustering in stage 1. In the process of time series data $X$ across 7 years, the LSTM layer with 64 units to retrieve 64-dimensional vectors of temporal features with career trend of each NBA player, combines with the clustering results of stage 1 to generate vectors including temporal features with 64 dimensions and types of career trend with extra dimensions, and the combination vectors process through a series of dense layers for career trend prediction. In the training phase, the 32 dimensions reduce to 16 dimensions gradually through the LSTM layers with ReLU activation, the Adam optimizer was used to avoid overfitting through regularization of early stop, and the training converges typically before reaching the maximum 100 epochs. Finally, the output layer generates 3 values showing the BPM score prediction of age 29, 30 and 31 years old. Hence, both patterns of individual performance and patterns of overall career trend were identified.

## IV. RESULTS AND DISCUSSIONS

### A. Prediction Performance Comparison of Machine Learning Models and Deep Learning Models

The proposed method was compared with common machine learning models and deep learning models for evaluating the prediction performance. In machine learning model, linear regression and ridge regression of basic models, random forest of ensemble learning, Support Vector Regression (SVR) of non-linear model, and a simple Last Value model for predicting the most recent BPM value were selected. In deep learning model, MLP, Gated Recurrent Unit (GRU), 1D CNN, a hybrid model of the CNN model and the LSTM model, and the bidirectional LSTM (BiLSTM) model were selected. In addition, a standard LSTM model was also selected without clustering for evaluation the clustering effect of the proposed method.

From the comparison results of prediction performance (Table II), the proposed method outperforms the other models (Test MAE=1.42, Test $R^2$=0.55), however some models with negative $R^2$ values may have poor performance than the simple mean prediction. The prediction performance of machine learning methods shows that the random forest performed best (Test MAE=1.48, Test $R^2$=0.49) and the linear regression model has worse performance with negative test $R^2$ values (Test $R^2$=-10.56). Compared to the prediction performance of random forest, the proposed method achieves 4.05% reduction in test MAE and 12.24% improvement in test $R^2$. The prediction performance of deep learning methods shows that the GRU model performed best (Test MAE=1.82, Test $R^2$=0.23) and the 1D CNN model has worse performance with negative test $R^2$ values (Test $R^2$=-0.20). Compared to the prediction performance of standard LSTM model without clustering (Test MAE=1.84, Test $R^2$=0.19), the proposed method achieves 22.83% reduction in test MAE and 189.47% improvement in test $R^2$ that reveals the clustering effect in the proposed method.

### B. Performance Analysis Across Player Categories

There are two types of career trend in the study (K=2). For evaluating the prediction performance of the proposed method, 222 NBA players were selected from the dataset and were divided into star player category (23 players, 10.4% of total number) and regular player category (199 players, 89.6% of total number) according to All-Star selection, award and team contribution. From the results of prediction performance (Table III), the proposed method improves significantly (Star player: 63.1%, regular player: 26.0%) than the standard LSTM model without clustering in both star player category and regular player category, and the standard LSTM model have negative $R^2$ values in both categories with poor performance.

TABLE II. PREDICTION PERFORMANCE COMPARISON

| Model Category | Model Name/ Test MAE/ Test $R^2$ | | |
|---|---|---|---|
| | *Model Name* | *Test MAE* | *Test $R^2$* |
| Machine Learning | Linear Regression | 6.66 | -10.56 |
| | Ridge Regression | 5.63 | -7.46 |
| | Random Forest | 1.48 | 0.49 |
| | SVR | 1.81 | 0.17 |
| | Last Value | 1.71 | 0.36 |
| Deep Learning | MLP | 1.93 | -0.01 |
| | GRU | 1.82 | 0.23 |
| | 1D CNN | 2.14 | -0.20 |
| | CNN + LSTM | 1.93 | 0.03 |
| | BiLSTM | 1.84 | 0.21 |
| | Standard LSTM | 1.84 | 0.19 |
| The Proposed Method | The proposed method | 1.42 | 0.55 |

TABLE III. PREDICTION PERFORMANCE COMPARISON OF STAR PLAYER AND REGULAR PLAYER

| Player Type | Model/ MA/ $R^2$/ Count | | | |
|---|---|---|---|---|
| | *Model* | *MAE* | *$R^2$* | *Count* |
| Star Player | Standard LSTM | 4.83 | -4.15 | 23 |
| Star Player | The proposed method | 1.78 | 0.23 | 23 |
| Regular Player | Standard LSTM | 1.96 | -0.07 | 199 |
| Regular Player | The proposed method | 1.45 | 0.39 | 199 |

TABLE IV. COMPARISON OF ACTUAL BPM VALUE AND PREDICTED BPM VALUE IN INDIVIDUAL PLAYER

| Player | Actual BPM | | | Standard LSTM | | | The proposed method | | |
|---|---|---|---|---|---|---|---|---|---|
| | *Age 29* | *Age 30* | *Age 31* | *Age 29* | *Age 30* | *Age 31* | *Age 29* | *Age 30* | *Age 31* |
| LeBron James | 8.80 | 7.10 | 9.00 | 0.30 | 0.24 | -0.29 | 7.00 | 5.89 | 5.55 |
| Stephen Curry | 7.70 | 6.60 | 3.90 | 0.04 | 0.09 | -0.57 | 6.35 | 5.43 | 4.71 |
| Michael Carter-Williams | -4.80 | -5.55 | -6.30 | 0.04 | 0.01 | -0.08 | -1.61 | -2.17 | -2.54 |
| Lance Stephenson | -1.93 | -2.37 | -2.80 | 0.08 | 0.09 | -0.17 | -1.59 | -1.81 | -2.25 |

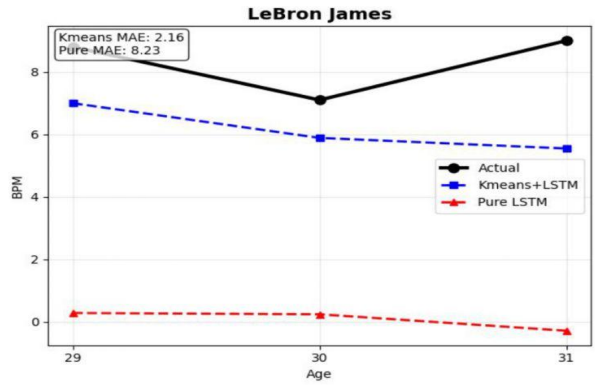
(a)
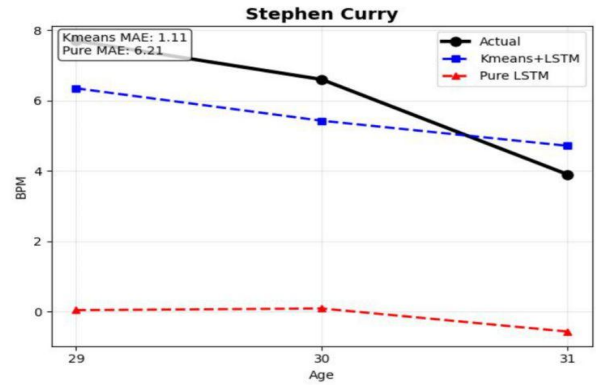
(b)
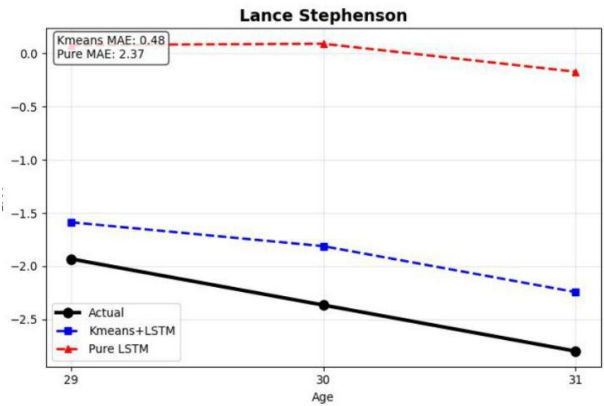
(c)
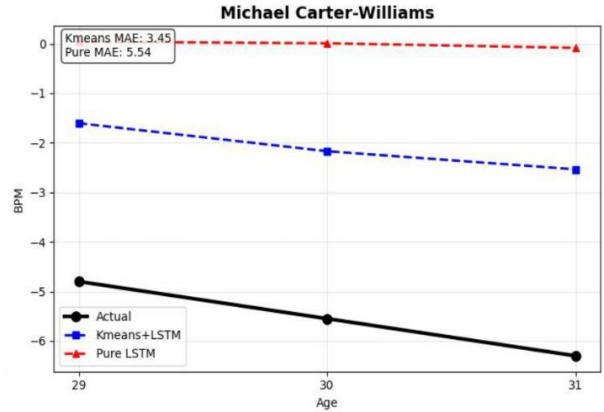
(d)

Figure 4. Career trend predictions of NBA players: star players of (a) LeBron James and (b) Stephen Curry, and regular players of (c) Lance Stephenson and (d) Michael Carter-Williams.

From the comparison results of BPM value among representative players in star player category and regular player category (Tabel IV), star players such as LeBron James and Stephen Curry have significant small prediction errors than the standard LSTM model without clustering, and regular players such as Michael Carter-Williams and Lance Stephenson also have significant career trend of individuals.

From the results of career trend prediction, there are obvious differences career trend among the star players (Figure 3 (a) and (b)) and regular players (Figure 3 (c) and (d)), and star players have greater variance and non-linear career trend. That also provides evidence of the need of clustering in the proposed method. The curves of the proposed method match the actual career trend of LeBron James (Start player), Stephen Curry (Start player) and Lance Stephenson (Regular player) closely, and the curves of the standard LSTM model without clustering has poor performance with near-zero values (Figure 3 (a), (b) and (c)). In the case of Michael Carter-Williams (Regular player) has challenges in the prediction results, however the proposed method performed better than the standard LSTM model without clustering (Figure 3 (d)). Therefore, the proposed method performed better in individuals of star player category than the individuals of regular player category for identifying complex patterns. In addition, from the aggregation of the individual patterns into the average career trend by player type, the propose method has high accuracy in career trend prediction both in star play category and regular player category (Figure 4). Moreover, from the comparison results of BPM performance in age 29 (Figure 5 (a)), 30 (Figure 5 (b)) and 31 (Figure 5 (c)) years old, the proposed method performed better than the standard LSTM model without clustering, and the prediction results close to the ideal prediction line in both star player category and regular player category. The standard LSTM model has poor performance with flat distribution away from the ideal prediction line that reveals only using the model fails to capture meaningful patterns in both player categories. Hence, the proposed method can predict career trend of different types of NBA players effectively.

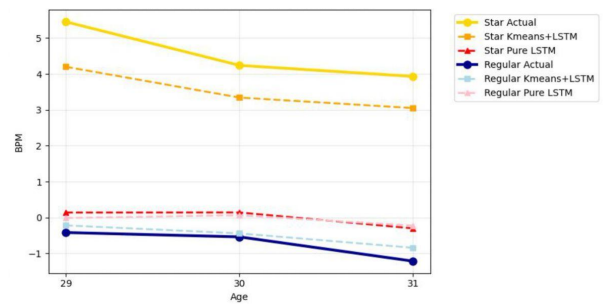

Figure 3. Average career trend by player category.

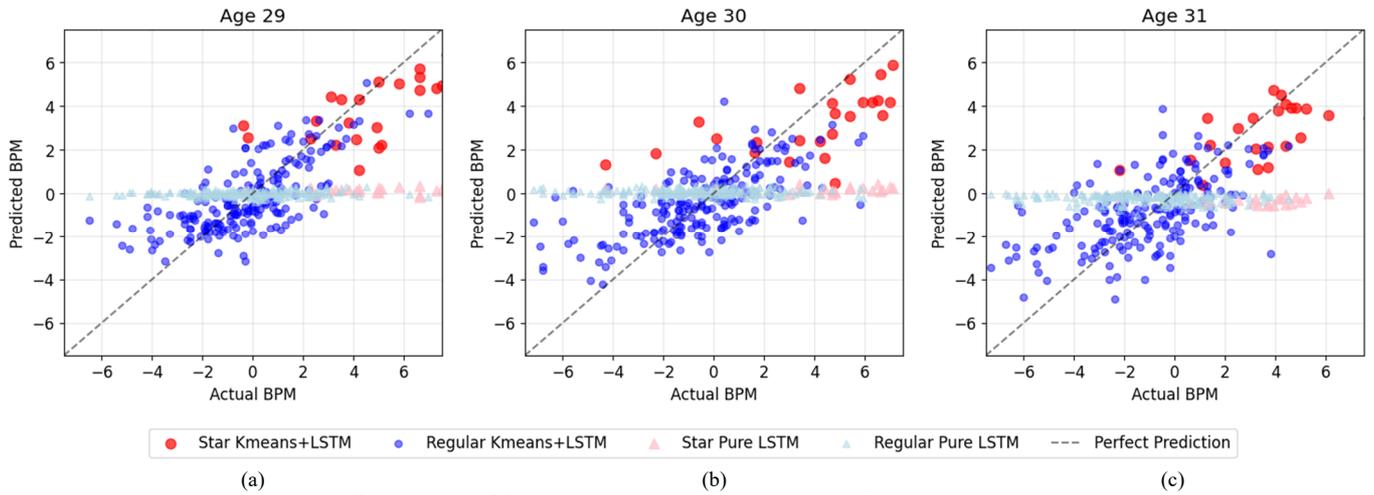

Figure 5. Comparison of BPM performance in different ages: (a) age 29, (b) age 30 and (c) age 31.

## V. CONCLUSION

A two-stage framework proposed combining autoencoder with K-means clustering and the LSTM model for NBA career forecasting. The autoencoder clustering helps to retrieve the diverse aging patterns of different types of NBA players, and reveals obvious different career trend between star player category and regular player category. The proposed method outperformed than the than the other methods with high accuracy in career trend prediction.

## ACKNOWLEDGMENTS

This work was partially supported by the National Science and Technology Council, Taiwan, under grants 114-2635-E-004 -001- and 114-2813-C-004-054-E.